\documentclass[conference,a4paper]{IEEEtran}
\IEEEoverridecommandlockouts

\usepackage[hidelinks]{hyperref}
\usepackage[cmex10]{amsmath}
\usepackage{amssymb,amsfonts}
\usepackage{dblfloatfix}
\usepackage{caption}
\usepackage[ruled,vlined]{algorithm2e}
\usepackage{graphicx}
\graphicspath{{Figures/PDF/}{Figures/PNG/}}
\usepackage{subcaption}
\usepackage{booktabs}
\usepackage{siunitx}
\usepackage[numbers,compress]{natbib}
\usepackage{texnames}
\usepackage{bm,bbm}
\usepackage{orcidlink}

\begin{document}

\title{\uppercase{Correlation-based band selection for hyperspectral image classification}}


\author{\IEEEauthorblockN{Dibyabha Deb}
\IEEEauthorblockA{\textit{Manipal Institute of Technology Bengaluru} \\
Manipal Academy of Higher Education, Manipal, India\\
dibyabha.deb@learner.manipal.edu
}
\and
\IEEEauthorblockN{ Ujjwal Verma}
\IEEEauthorblockA{\textit{Manipal Institute of Technology} \\
Manipal Academy of Higher Education, Manipal, India\\
ujjwal.verma@manipal.edu
}
}

\maketitle
\begin{abstract}
Hyperspectral images offer extensive spectral information about ground objects across multiple spectral bands. However, the large volume of data can pose challenges during processing. Typically, adjacent bands in hyperspectral data are highly correlated, leading to the use of only a few selected bands for various applications. In this work, we present a correlation-based band selection approach for hyperspectral image classification. Our approach calculates the average correlation between bands using correlation coefficients to identify the relationships among different bands. Afterward, we select a subset of bands by analyzing the average correlation and applying a threshold-based method. This allows us to isolate and retain bands that exhibit lower inter-band dependencies, ensuring that the selected bands provide diverse and non-redundant information. We evaluate our proposed approach on two standard benchmark datasets: Pavia University (PA) and Salinas Valley (SA), focusing on image classification tasks.  The experimental results demonstrate that our method performs competitively with other standard band selection approaches.
\end{abstract}

\begin{IEEEkeywords}
	Hyperspectral Image, Image Classification, Band Selection, Correlation
\end{IEEEkeywords}

\section{Introduction}

Hyper-Spectral Imaging (HSI) is an important component of remote sensing and is attributed to the extraction of comprehensive information. The Hyper-Spectral image (HS) is a 3D data cube that consists of numerous detailed spectral and spatial information, covering a wide range of electromagnetic spectrum (\(0.4\mu m\) to \(2.5\mu m\)). HS can be represented as \(\mathbf{X} \in \mathbb{R}^{W \times H \times N}\), where \(W\) and \(H\) represent the width and height of the image, and \(N\) represents the number of spectral bands, and \(\mathbf{X}\) represents the image. These images, from different regions covering a specific area, are beyond human vision (visible range \(0.4 \mu m\) to \(0.7 \mu m\)) and have contiguous spectral bands.\\
\hspace*{1em}HSI has been used in several real world applications, including, but not limited to, environmental monitoring, urban and climate studies, agriculture analysis, and military surveillance \cite{ardouin2007demonstration} \cite{article} \cite{gi-6-169-2017}. Furthermore, HSI is also used in examining celestial bodies, analyzing Earth's surface, and identifying minerals \cite{10.1117/12.830284} \cite{7835190}.\\
\hspace*{1em}Despite the presence of hundreds of bands in a HS, each individual band may not provide distinct or significant information. Oftentimes, the adjacent bands of the selected bands contain more valuable information as they capture subtle spectral differences, which can lead to more accurate analysis for a particular task such as classification, object detection, dimensionality reduction, and segmentation. The inclusion of adjacent bands of the initially selected bands can enhance the accuracy of the analysis, but there is a need to maintain a trade-off between the number of bands selected and performance. Specifically, maintaining a trade-off between the number of bands and performance is crucial as selection of too many bands can lead to poor analysis, increase the computational cost, and introduce redundant information.\\
\hspace*{1em}Due to the high dimensionality of HS, it becomes essential to employ dimensional reduction or band selection techniques to select the relevant bands and reduce the dimension for further analysis of HS. These techniques help in mitigating the curse of dimensionality or Hughes phenomenon, a phenomenon where the efficiency and effectiveness deteriorate as dimension of data increases after the optimal features are selected/extracted and leads to overfitting and computational complexity \cite{1054102}. Therefore, a typical analysis of hyperspectral images starts with the selection of a few hyperspectral bands to reduce spectral redundancy and computational costs \cite{8899002}. There are several methods for band selection, such as those based on transforming the original data into another feature space (feature extraction). Other approaches include ranking, clustering, and sparsity-based methods for selecting the optimal bands in hyperspectral data \cite{8738051} \cite{9324974} \cite{8288979} \cite{pcaarticle} \cite{airborne} \cite{wave}.\\
\hspace*{1em}In HS classification task, the determination of relevant bands is a challenging problem. The feature selection methods suffers from certain drawbacks such as local minima problem and information loss, where potential or relevant valuable information is discarded along with redundant data. Furthermore, such methods can also be sensitive to initialization and parameter fine-tuning. Feature extraction methods can mitigate the disadvantages of feature selection to a certain extent by revealing the most vital information. They can be further divided into supervised and unsupervised methods. In supervised methods, a priori knowledge is required for computation, whereas unsupervised methods can extract features without a priori knowledge \cite{pcaarticle}.\\
\hspace*{1em}In this article, we have explored the use of Correlation Coefficient (CC) as a primary feature extraction method to mitigate the challenges mentioned above due to high dimensional data. The proposed method considers, initially, all \textit{N} bands of the HS and extracts relevant information while maintaining the integrity of the data. By analyzing the correlation between the spectral bands, the method identifies and retains a subset of the \textit{least} correlated bands using a threshold-based selection approach. This threshold-based selection is a straightforward determination of bands that contribute unique information. This band reduction process effectively decreases the dimensionality of the dataset, reduces computational requirements, and enhances processing speed without significant loss of information. After reducing the band, we utilized a Support Vector Machine (SVM) classifier to assess its performance on the hyperspectral (HS) classification task using publicly available benchmark datasets.

\section{Methodology}
The proposed work computes band correlation using Correlation Coefficient (CC) to determine the relationships between different spectral bands of the datasets. This metric was chosen to quantify the inter-band dependencies of the benchmark datasets. CC quantifies the linear relationship between two variables (or bands in our study) and describes how they move in relation to each other. It ranges from -1 to 1, where -1 indicates a perfect negative correlation (as one variable increases, the other decreases), and 1 indicates a perfect positive correlation (as one variable increases, the other also increases).
\begin{table*}[htbp]
    \centering
    \caption{COMPARATIVE EVALUATION OF PCA \cite{pcaarticle}, SB \cite{4656481} AND ABC BASED FEATURE EXTRACTION (PROPOSED) METHODS ON SA DATASET}
    \label{tab:SA}
    \begin{tabular}{|c|c|c|c|c|c|c|c|c|c|}
        \hline
        CLASS & \multicolumn{3}{c|}{PCA} & \multicolumn{3}{c|}{SB} & \multicolumn{3}{c|}{PROPOSED} \\
        \cline{2-10}
        & PRECISION & RECALL & F1 & PRECISION & RECALL & F1 & PRECISION & RECALL & F1 \\
        \hline
        1 & 1 & 0.98 & 0.99 & 1 & 0.98 & 0.99 & 1 & 0.98 & 0.99 \\
        \hline
        2 & 0.99 & 1 & 0.99 & 0.99 & 1 & 0.99 & 0.99 & 1 & 1 \\
        \hline
        3 & 0.89 & 0.98 & 0.93 & 0.9 & 0.98 & 0.94 & 0.91 & 0.98 & 0.95 \\
        \hline
        4 & 0.98 & 1 & 0.99 & 0.98 & 1 & 0.99 & 0.99 & 1 & 0.99 \\
        \hline
        5 & 0.99 & 0.97 & 0.98 & 0.99 & 0.96 & 0.97 & 0.98 & 0.95 & 0.97 \\
        \hline
        6 & 1 & 1 & 1 & 1 & 1 & 1 & 1 & 1 & 1 \\
        \hline
        7 & 0.97 & 0.99 & 0.98 & 1 & 0.99 & 1 & 1 & 0.99 & 1 \\
        \hline
        8 & 0.73 & 0.88 & 0.8 & 0.74 & 0.92 & 0.82 & 0.73 & 0.91 & 0.81 \\
        \hline
        9 & 0.99 & 0.99 & 0.99 & 0.99 & 1 & 1 & 0.97 & 0.98 & 0.97 \\
        \hline
        10 & 0.9 & 0.89 & 0.9 & 0.91 & 0.93 & 0.92 & 0.94 & 0.92 & 0.93 \\
        \hline
        11 & 0.93 & 0.89 & 0.91 & 0.95 & 0.9 & 0.92 & 0.82 & 0.78 & 0.8 \\
        \hline
        12 & 0.96 & 1 & 0.98 & 0.96 & 1 & 0.98 & 0.93 & 1 & 0.96 \\
        \hline
        13 & 0.95 & 0.98 & 0.96 & 0.94 & 0.98 & 0.96 & 0.94 & 0.98 & 0.96 \\
        \hline
        14 & 0.98 & 0.92 & 0.95 & 0.97 & 0.93 & 0.95 & 0.98 & 0.93 & 0.95 \\
        \hline
        15 & 0.77 & 0.5 & 0.61 & 0.83 & 0.51 & 0.64 & 0.79 & 0.5 & 0.61 \\
        \hline
        16 & 0.98 & 0.92 & 0.95 & 1 & 0.98 & 0.99 & 0.99 & 0.98 & 0.99 \\
        \hline
        OA & \multicolumn{3}{|c|}{89.08} & \multicolumn{3}{|c|}{90.41} & \multicolumn{3}{|c|}{89.33} \\
        \hline
        KAPPA & \multicolumn{3}{|c|}{0.88} & \multicolumn{3}{|c|}{0.89} & \multicolumn{3}{|c|}{0.88} \\
        \hline
    \end{tabular}
\end{table*}

\begin{table*}[htbp]
    \centering
    \caption{COMPARATIVE EVALUATION OF PCA \cite{pcaarticle}, SB \cite{4656481} AND ABC BASED FEATURE EXTRACTION (PROPOSED) METHODS ON PA DATASET}
    \label{tab:PA}
    \begin{tabular}{|c|c|c|c|c|c|c|c|c|c|}
        \hline
        CLASS & \multicolumn{3}{c|}{PCA} & \multicolumn{3}{c|}{SB} & \multicolumn{3}{c|}{PROPOSED} \\
        \cline{2-10}
        & PRECISION & RECALL & F1 & PRECISION & RECALL & F1 & PRECISION & RECALL & F1 \\
        \hline
        1 & 0.78 & 0.93 & 0.85 & 0.91 & 0.93 & 0.92 & 0.87 & 0.91 & 0.89 \\
        \hline
        2 & 0.81 & 0.99 & 0.89 & 0.91 & 0.99 & 0.95 & 0.91 & 0.99 & 0.95 \\
        \hline
        3 & 0.63 & 0.37 & 0.46 & 0.84 & 0.67 & 0.75 & 0.79 & 0.55 & 0.65 \\
        \hline
        4 & 0.95 & 0.87 & 0.91 & 0.97 & 0.93 & 0.95 & 0.97 & 0.92 & 0.94 \\
        \hline
        5 & 1 & 1 & 1 & 1 & 1 & 1 & 1 & 1 & 1 \\
        \hline
        6 & 0.94 & 0.23 & 0.37 & 0.96 & 0.69 & 0.8 & 0.95 & 0.67 & 0.79 \\
        \hline
        7 & 0.91 & 0.09 & 0.16 & 0.9 & 0.74 & 0.82 & 0.86 & 0.65 & 0.74 \\
        \hline
        8 & 0.74 & 0.86 & 0.8 & 0.82 & 0.9 & 0.86 & 0.78 & 0.9 & 0.84 \\
        \hline
        9 & 1 & 1 & 1 & 1 & 1 & 1 & 1 & 1 & 1 \\
        \hline
        OA & \multicolumn{3}{|c|}{81.45} & \multicolumn{3}{c|}{91} & \multicolumn{3}{c|}{89.56} \\
        \hline
        KAPPA & \multicolumn{3}{|c|}{0.74} & \multicolumn{3}{c|}{0.88} & \multicolumn{3}{c|}{0.86} \\
        \hline
    \end{tabular}
    
\end{table*}

Mathematically, CC can be represented as \(r_{XY}\) where \(X\) and \(Y\) are the two variables (or bands) and can be formulated as
\begin{equation} 
r_{XY} = \frac{\sum_{i=1}^{n} (X_i - \bar{X})(Y_i - \bar{Y})}{\sqrt{\sum_{i=1}^{n} (X_i - \bar{X})^2\sum_{i=1}^{n} (Y_i - \bar{Y})^2}} \label{eq:correlation_coefficient} 
\end{equation} where 
\(X_i\) and \(Y_i\) are the individual samples and \(\bar{X}\) and \(\bar{Y}\) are the mean values of variables \(X\) and \(Y\) respectively, and \(n\) is the number of samples (pixels) in each variable (or band) after pre-processing. The band correlation data was stored as a matrix for future access.\\
\hspace*{1em}  Subsequently, Average Band Correlation (ABC) is computed. The ABC for band \textit{i} is defined as the mean of the absolute correlation of band \textit{i} (\(B_i\)) with band \textit{j} (\(B_j\)), where \textit{j} varies from 1 to N and \(j \neq i\). 
\begin{equation} 
\text{ABC}_i = \frac{1}{N-1} \sum_{j=1, j \neq i}^{N} \left| r_{B_i, B_j} \right|
\end{equation}
where \({r}(B_i, B_j)\) represents the correlation coefficient between \(B_i\) and \(B_j\) , and \(|\cdot|\) denotes the absolute value. This process is repeated for each \(B_i\), where \(i \in \{1,\ldots,N\}\)\\
\hspace*{1em} We experimentally set a threshold of 0.65 for the average band correlation (ABC). Bands with ABC less than the threshold were selected and, these selected bands were then extracted from the datasets. This approach allowed us to isolate and retain bands that exhibited lower inter-band dependencies, ensuring that the retained bands provided diverse and non-redundant information.


In contrast to previous studies that utilized a grouping strategy \cite{group1} \cite{group2}, we adopted a threshold-based approach. This method offers a straightforward criterion for selecting bands. By establishing a threshold, our goal is to decrease the total number of bands used for a specific task while retaining those bands that demonstrate lower correlations.

\begin{figure*}[htbp] 
\centering 
\begin{subfigure}{0.22\textwidth} 
\centering 
\includegraphics[width=\linewidth]{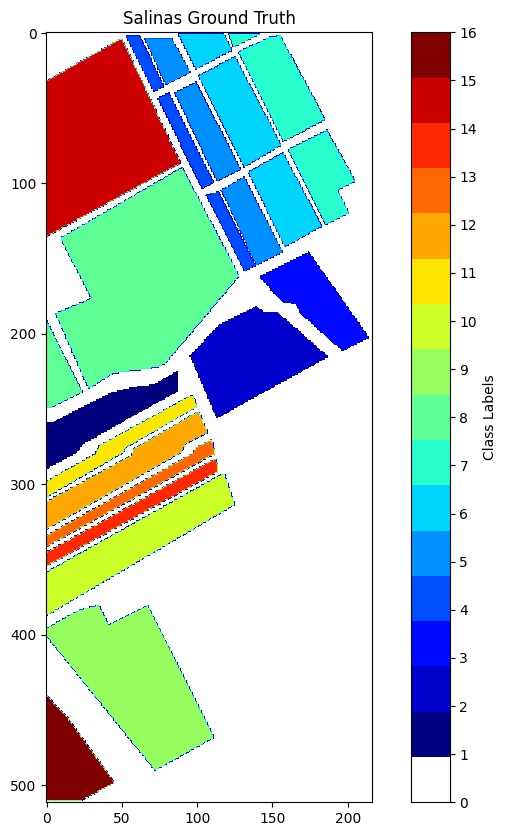} 
\caption{Ground Truth (SA)} 
\label{fig:sagt} 
\end{subfigure} 
\begin{subfigure}{0.22\textwidth} 
\centering 
\includegraphics[width=\linewidth]{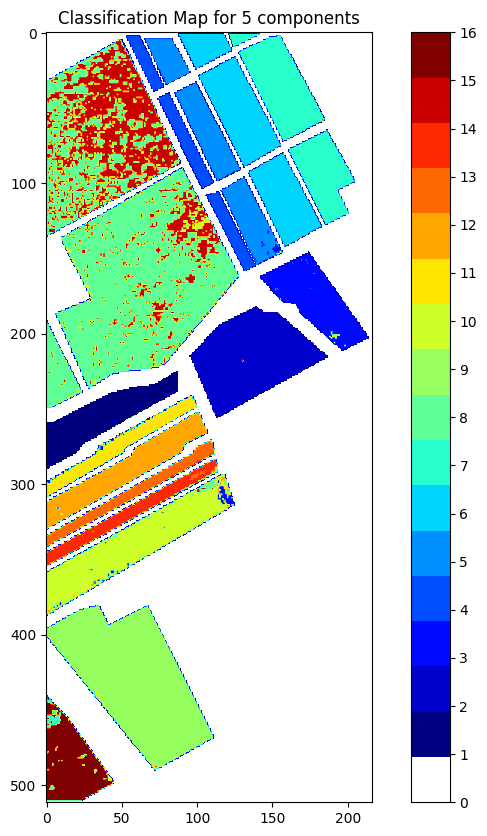} 
\caption{PCA}%
\label{fig:sapca} 
\end{subfigure} 
\begin{subfigure}{0.22\textwidth} 
\centering 
\includegraphics[width=\linewidth]{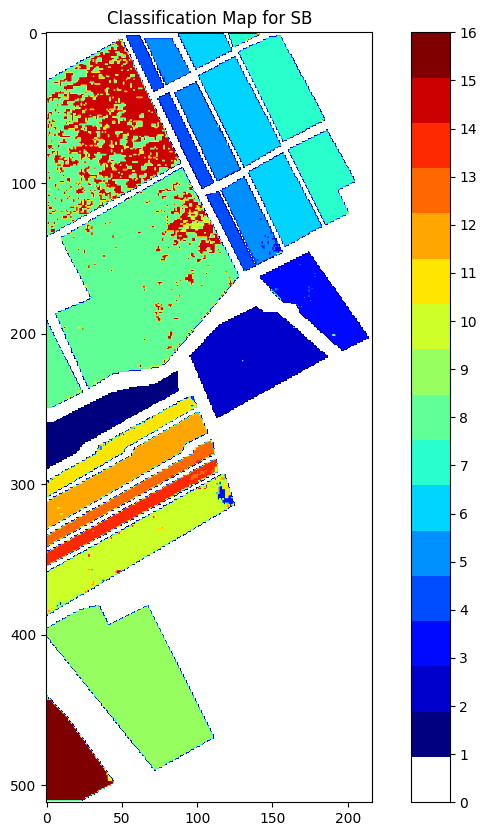} 
\caption{SB} 
\label{fig:sasb} 
\end{subfigure}
\begin{subfigure}{0.22\textwidth} 
\centering 
\includegraphics[width=\linewidth]{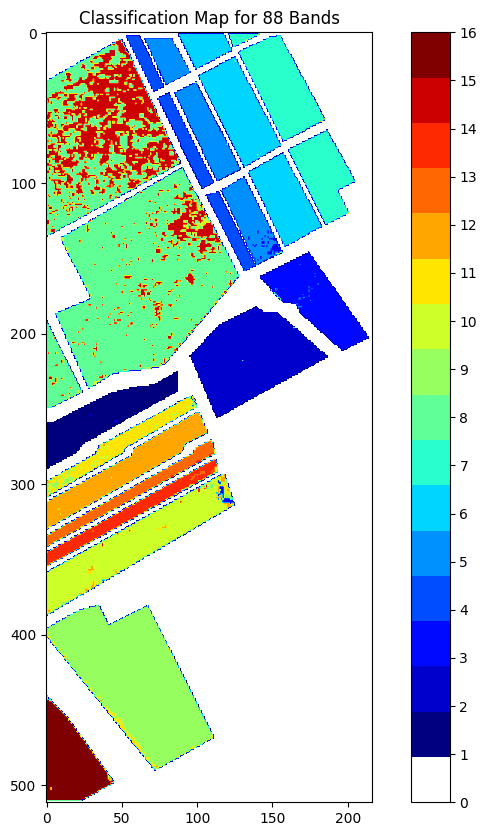} 
\caption{Proposed}%
\label{fig:sacc} 
\end{subfigure}
\begin{subfigure}[b]{0.22\textwidth} 
\centering 
\includegraphics[width=\linewidth]{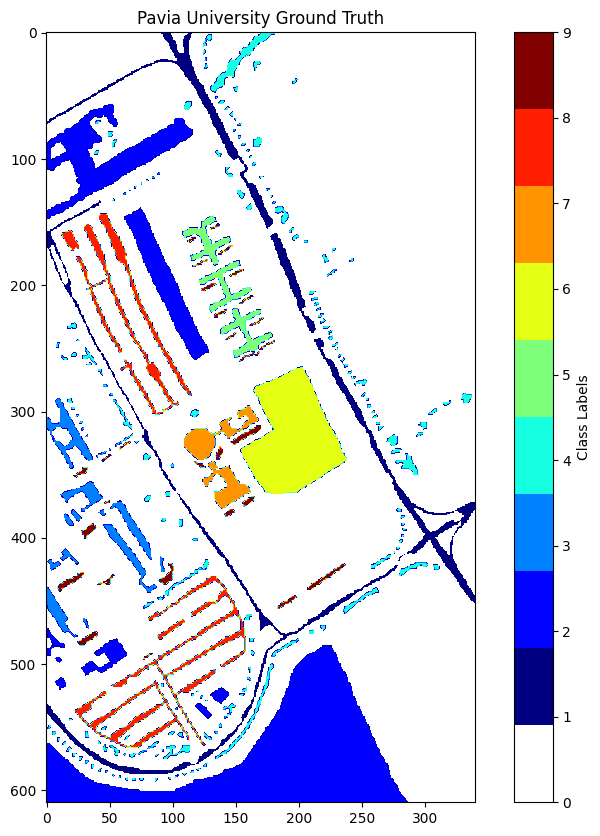} 
\caption{Ground Truth (PA)}
\label{fig:pagt} 
\end{subfigure} 
\begin{subfigure}[b]{0.22\textwidth} 
\centering 
\includegraphics[width=\linewidth]{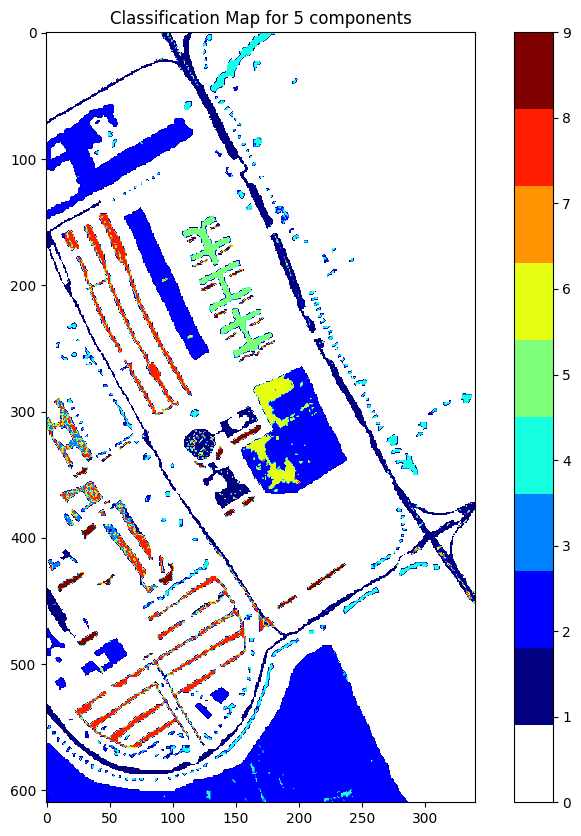} 
\caption{PCA} 
\label{fig:papca} 
\end{subfigure} 
\begin{subfigure}[b]{0.22\textwidth} 
\centering 
\includegraphics[width=\linewidth]{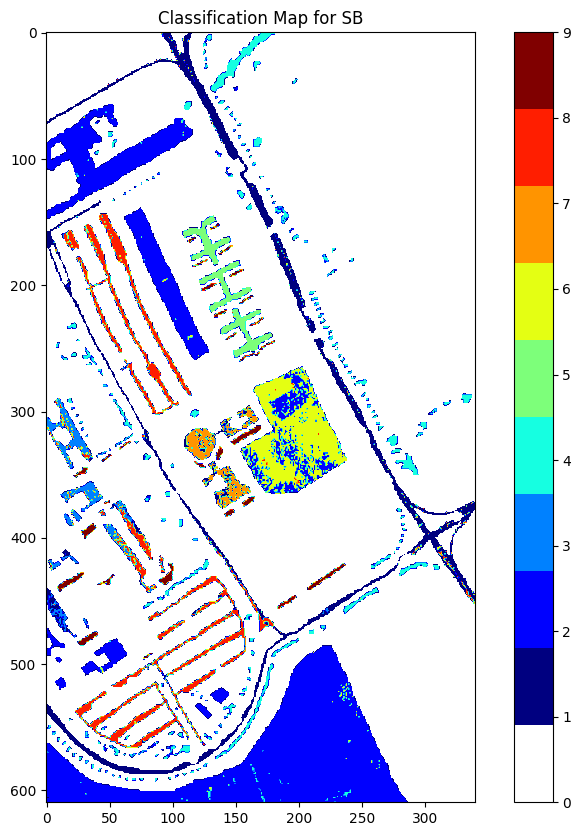} 
\caption{SB} 
\label{fig:pasb}
\end{subfigure}
\begin{subfigure}[b]{0.22\textwidth} 
\centering 
\includegraphics[width=\linewidth]{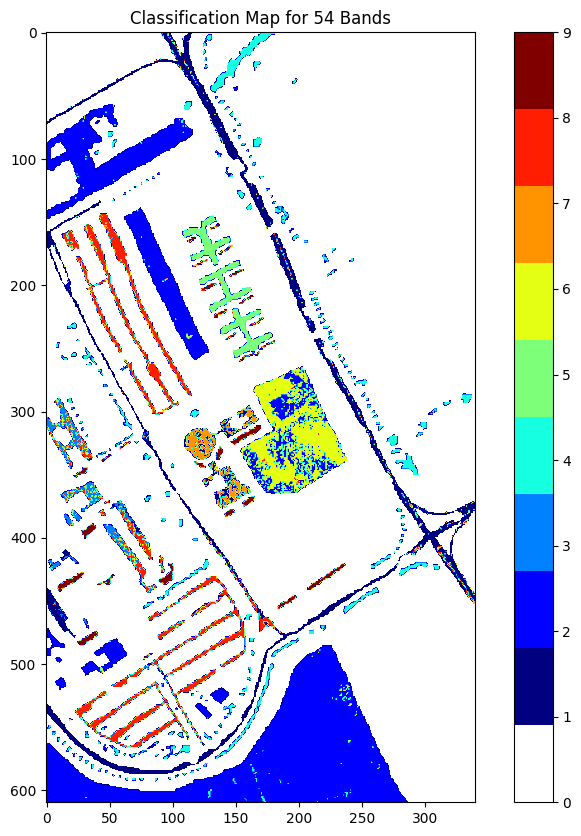} 
\caption{Proposed} 
\label{fig:pacc}
\end{subfigure}
\caption{Comparison of Ground Truth, PCA, SB and Proposed method for SA and PA datasets. (a) and (e) are Ground Truth maps, (b) and (f) are Classification maps using PCA method, (c) and (g) are Classification maps using SB method and, (d) and (h) are Classification maps using Proposed method on SA and PA datasets, respectively.}
\label{fig:comparison} 
\end{figure*}

\section{Experiment and Results}
In this study, we utilized two publicly available datasets - Pavia University (PA) and Salinas Valley (SA) - to evaluate the performance of the proposed band selection approach on image classification \cite{data}. The details related to these datasets are provided in Table \ref{table:1}.
\begin{table}[!ht] 
\centering 
\caption{\\INFORMATION RELATED TO THE DATASETS USED}
\setlength{\tabcolsep}{0.35\tabcolsep}
\label{table:1} 
\begin{tabular}{|c|c|c|c|c|c|}
    \hline
    \textbf{DATASET} & \textbf{WAVELENGTH} & \textbf{SPATIAL SIZE} & \textbf{BANDS USED} & \textbf{CLASSES} \\
    \hline
    PA & 0.43 - 0.86$\mu$m & 610 $\times$ 340 & 103 & 9 \\ 
    \hline
    SA & 0.4 - 2.5$\mu$m & 512 $\times$ 217 & 204 & 16 \\ 
    \hline
    \end{tabular} 
\end{table}

The SA dataset was captured by the AVIRIS Sensor in aerial mode over the Salinas valley in California, originally consisting of 224 bands. However, 20 bands were excluded due to their coverage of water absorption region and low Signal-to-Noise ratio (SNR). The PA dataset was captured by the ROSIS sensor from Pavia in Northern Italy. After the removal of certain broken bands, 103 bands were treated for research analysis. As part of the data pre-processing step, all pixels were standardized before any further analysis and additionally, any pixels containing the background class were removed to ensure the integrity of the subsequent analysis.\\
\hspace*{1em}To facilitate the training and evaluation of the classification model, we divided the pixels of the images from both the datasets into training and testing sets, with 70\% of the pixels allocated for training and 30\% reserved for testing. This train-test split is crucial because it ensures that the model is trained on a subset of the data and then evaluated on an unseen subset, thereby providing a reliable measure of the model's generalization ability.\\
\hspace*{1em}We used Support Vector Machine (SVM) classifier to classify the pixels into different classes. SVM operates by finding the optimal hyperplane that separates the data points of different classes in a high-dimensional space. This is achieved by maximizing the margin between the closest data points of the classes known as support vectors. During the training, the SVM model was trained on the training set pixels and was subsequently, evaluated on the testing set pixels to check the performance of the model. The evaluation metrics used to assess the performance of the SVM classifier included class-wise precision, recall, and F1 score, as well as overall accuracy (OA) and the kappa coefficient (KAPPA).\\
\hspace*{1em}To evaluate the effectiveness of our proposed method, we compared it with two different methods - Principal Component Analysis (PCA) which deals with dimensionality reduction \cite{pcaarticle} and a similarity based unsupervised approach (SB) which deals with band selection \cite{4656481}. PCA based feature extraction was employed on both the datasets, and we retained the upper five principal components with largest eigenvalues. This approach explained 99.51\% and 98.83\% cumulative variance of PA and SA datasets, respectively.

For similarity based unsupervised approach (SB) \cite{4656481}, we selected the same number of bands as computed from our approach to ensure a fair comparison. The proposed approach with a threshold of 0.65 resulted in 88 (for SA) and 54 (for PA) bands. This allowed a direct comparison of the efficacy of the methods under consistent conditions, thereby providing a comprehensive assessment.

The performance metrics, including precision, recall, and F1 score for each class, as well as OA and KAPPA are summarized in Table \ref{tab:SA} and Table \ref{tab:PA} for the SA and PA datasets, respectively. The comparison between the ground truth maps and the classification maps obtained from the baseline and the proposed methods for the SA and PA datasets is shown in Fig \ref{fig:comparison}.

The comparative analysis presented in Tables \ref{tab:SA} and \ref{tab:PA} demonstrates that the proposed method achieved comparable results to the two different approaches namely dimensionality reduction using PCA and band selection methods using unsupervised learning (SB). In Table \ref{tab:SA}, all three methods - PCA-based method, SB-based method and proposed method - exhibit similar overall accuracies. However, in Table \ref{tab:PA}, the proposed method surpasses the PCA-based extraction method in terms of OA and achieves comparable accuracy to SB-based method. This improvement is attributed to the lower class-wise precision, recall and f1 in classes 3, 6, and 7 observed with PCA analysis, compared to SB-based and proposed methods.\\
\hspace*{1em}Upon closer examination of Fig \ref{fig:comparison}, it becomes evident that there are discrepancies between ground truths and classification maps obtained using different methods. Specifically, Fig \ref{fig:sagt} and Fig \ref{fig:sapca} reveals that there are discrepancies in the classification map obtained using PCA and ground truth for class label 16 of SA dataset. Similarly, there exists discrepancies between the ground truth and classification map obtained using PCA-based method when we look closer at Fig \ref{fig:pagt} and Fig \ref{fig:papca}. It shows that the classes 3, 6, and 7 of PA dataset are not properly classified. Our proposed method and SB-based method are quite comparable in both the datasets.

\section{Conclusion and Future work}
This paper proposes a correlation-based approach for selecting optimal bands for hyperspectral image classification. The adjacent bands in a hyperspectral image are correlated, and this correlation information is exploited to select an optimal number of bands. Specifically, the proposed approach calculates the average correlation between bands using correlation coefficients. A subset of bands is then selected by thresholding this average band correlation. The proposed method is evaluated on an image classification task on two benchmark datasets : Pavia University (PA) and Salinas Valley (SA). The proposed approach obtained an overall accuracy of 89.56 on PA as compared to 91 for a PCA-based approach and 81.41 using the unsupervised band selection approach . In addition, an overall accuracy of 89.93 on the SA dataset was obtained using the proposed approach as compared to 90.41 using a PCA-based approach and 89.08 using the unsupervised band selection approach. The experimental results demonstrate that the proposed straightforward approach of selecting optimal bands performs competitively with another standard approach for band selection.\\
\hspace*{1em}In future, the proposed method will be further developed to eliminate the need for determining a threshold. In fact, identifying an optimal threshold is a challenging task. Besides, additional information on theoretic metrics will be explored to further improve the performance of the proposed approach in selecting the optimal bands.  Furthermore, we intend to develop a metric that can quantitatively estimate information loss during the band reduction process. This metric will provide valuable insights into the trade-offs between dimensionality reduction and information retention, enabling more informed decisions when applying our method to different datasets.

\section{Data Availability}
The code utilized for the analyzes mentioned in this paper is publicly available on GitHub. To access the repository, visit the link (\href{https://github.com/Dibyabha/hsi-cc}{GitHub Link}). The repository includes detailed descriptions and instructions for replicating the results presented in the paper. The datasets used in this paper can be downloaded from this website \cite{data}. 
\small
\bibliographystyle{IEEEtranN}
\bibliography{references}

\end{document}